\documentclass[preprint,12pt]{elsarticle}

\usepackage{amssymb}
\usepackage{amsthm}
\usepackage{lineno}
\usepackage{bm}
\usepackage[colorlinks=true, pdfauthor=author]{hyperref}
\usepackage{wrapfig}
\usepackage{booktabs} 
\usepackage{algorithm}
\usepackage{algorithmic}
\usepackage{graphicx}
\usepackage{amsmath}
\usepackage{amssymb}
\usepackage{amsfonts}
\usepackage{bbding}
\usepackage[table]{xcolor}
\usepackage{multirow}
\usepackage{graphicx}
\usepackage{mathrsfs}
\usepackage{colortbl}
\usepackage{subfigure}
\usepackage{indentfirst}
\usepackage{ulem}
\usepackage{url}
\usepackage{verbatim}
\usepackage{rotating}
\usepackage{adjustbox}
\usepackage{wrapfig}
\usepackage{setspace}

\graphicspath{{./images/}}

\journal{arXiv}
\begin{document}

\begin{frontmatter}



\title{Improving GBDT Performance on Imbalanced Datasets: An Empirical Study of Class-Balanced Loss Functions}

\author[first]{Jiaqi Luo}
\ead{jluo@fields.utoronto.ca}
\author[second]{Yuan Yuan}
\ead{y.yuan@dukekunshan.edu.cn}
\author[second]{Shixin Xu \corref{cor1}}
\cortext[cor1]{Corresponding author}
\ead{shixin.xu@dukekunshan.edu.cn}
\affiliation[first]{organization={The Fields Institute for Research in Mathematical Sciences},
            addressline={222 College Street}, 
            city={Toronto},
            postcode={M5T3J1}, 
            state={Ontario},
            country={Canada}}
\affiliation[second]{organization={Data Science Research Center, Duke Kunshan University},
            addressline={No.8 Duke Avenue}, 
            city={Kunshan},
            postcode={215000}, 
            state={Jiangsu Province},
            country={China}}

\begin{abstract}

Class imbalance remains a significant challenge in machine learning, particularly for tabular data classification tasks. While Gradient Boosting Decision Trees (GBDT) models have proven highly effective for such tasks, their performance can be compromised when dealing with imbalanced datasets.
This paper presents the first comprehensive study on adapting class-balanced loss functions to three GBDT algorithms across various tabular classification tasks, including binary, multi-class, and multi-label classification.
We conduct extensive experiments on multiple datasets to evaluate the impact of class-balanced losses on different GBDT models, establishing a valuable benchmark.
Our results demonstrate the potential of class-balanced loss functions to enhance GBDT performance on imbalanced datasets, offering a robust approach for practitioners facing class imbalance challenges in real-world applications. 
Additionally, we introduce a Python package that facilitates the integration of class-balanced loss functions into GBDT workflows, making these advanced techniques accessible to a wider audience.
The code is available at \url{https://github.com/Luojiaqimath/ClassbalancedLoss4GBDT}

\end{abstract}




\end{frontmatter}


\section{Introduction}
\label{s:intro}
Gradient Boosting Decision Tree (GBDT) \cite{chen2016xgboost, ke2017lightgbm, iosipoi2022sketchboost, luo2023trboost} has emerged as a powerful and versatile machine learning technique, widely adopted across various domains due to their exceptional predictive performance. However, like many machine learning algorithms, GBDT faces challenges when dealing with imbalanced datasets.

Class imbalance is a persistent issue in many real-world applications, such as fraud detection \cite{zhu2020optimizing}, medical diagnosis \cite{santos2022decision}, and fault diagnosis \cite{zhang2018imbalanced}. It poses significant challenges to machine learning algorithms, leading to poor performance, particularly in predicting the minority class.
Various strategies have been prompted to address this challenge, including sampling techniques and algorithm modifications \cite{he2009learning, fernandez2018learning}. While these methods have shown promise, the exploration of class-balanced losses, particularly in multi-label classification, has received comparatively little attention.

This paper presents the first comprehensive study on adapting class-balanced loss functions to GBDT algorithms across various tabular classification tasks, including binary, multi-class, and multi-label classification. We conduct extensive experiments on multiple datasets spanning diverse classification tasks, rigorously evaluating the performance of class-balanced losses within different GBDT models. Our thorough results demonstrate the effectiveness of these loss functions in mitigating class imbalance issues in tree-based ensemble methods.

To facilitate broader adoption and further research, we have developed and released a Python package \footnote{\url{https://pypi.org/project/gbdtCBL/0.1/}} that implements the losses. This tool allows for easy integration of class-balanced loss functions into existing GBDT workflows, making it accessible to both academic researchers and industry practitioners.

By bridging the gap between class-balanced loss functions and GBDT algorithms, our work aims to enhance the performance of these powerful models on imbalanced datasets, potentially opening new avenues for their application in challenging real-world scenarios.

\section{Related work}
\label{s:rel}
\subsection{GBDT for imbalanced classification}

In the context of GBDT for imbalanced learning, methodologies can be roughly classified into two distinct categories: data-level methods and algorithm-level methods.

Data-level methods use sampling techniques to balance the class distribution before applying GBDT. The popular methods are over-sampling \cite{chawla2002smote, kovacs2019empirical}, under-sampling \cite{yen2009cluster, batista2004study, mani2003knn} and hybrid-sampling \cite{ramentol2012smote, stefanowski2008selective}. However, these techniques have significant drawbacks. Over-sampling can introduce redundant data, while under-sampling might discard valuable information. Hybrid-sampling, which combines both methods, also risks adding meaningless samples to the dataset.

On the algorithmic side, existing learners are modified to reduce bias toward majority classes. This includes cost-sensitive learning \cite{xu2020imbalanced, liu2022focal, liu2022predicting} and loss function modification \cite{wang2020imbalance, luo2023robust, tian2023synergetic, mushava2024flexible}. 
In cost-sensitive learning, higher costs are assigned to misclassifying minority class samples, compelling the model to focus on them. However, determining the appropriate cost values is challenging due to various influencing factors. Another approach involves using class-balanced loss functions, which are straightforward to implement and leverage the efficiency of modern GBDT models. Despite their benefits, these methods are often limited by their reliance on convexity and easy Hessian implementation.

\subsection{Class-balanced loss}
The key idea behind class-balanced losses is to adjust the loss function to give more weight to underrepresented classes.
Most class-balanced loss functions originate from deep learning because of its flexibility and automatic differentiation capabilities. Deep learning frameworks easily adapt these loss functions, allowing for more precise adjustments and improved handling of class imbalances.

Weighted cross-entropy \cite{sun2009classification} assigns greater importance to minority samples, prompting the model to give extra attention to classifying these instances correctly. Focal loss \cite{lin2017focal} alters the standard cross-entropy loss by introducing a factor that diminishes the loss for well-classified examples while emphasizing the harder, misclassified ones, thus guiding the model to focus on challenging cases.

Class-balanced cross-entropy \cite{cui2019class} incorporates the effective number of samples into the loss function to enhance performance by accounting for data overlap. In label-distribution-aware margin loss \cite{cao2019learning}, the classification margin for each class is adjusted based on its sample size, which allows for larger margins for minority classes, thereby boosting their performance without sacrificing accuracy for majority classes. Influence-balanced loss \cite{park2021influence} tackles imbalanced data by modifying the training process to adjust the loss function according to each sample's impact.
Dynamically Weighted Balanced Loss \cite{fernando2021dynamically} adapts class weights during training by considering class frequency and model confidence. This approach addresses class imbalance and improves confidence in deep neural networks. 

Distribution-balanced loss \cite{wu2020distribution}, designed for multi-label imbalanced classification, accounts for label co-occurrence and incorporates negative-tolerant regularization to mitigate the dominance of negative labels.
Asymmetric loss \cite{ridnik2021asymmetric} refines focal loss by shifting the probability for negative parts and setting different focusing parameters. It’s a simple, effective, and easy-to-implement method.

Equalized Focal Loss \cite{li2022equalized}, tailored for one-stage detectors in object detection, tackles varying degrees of positive-negative imbalance found in long-tailed data distributions. Equalization Loss \cite{tan2020equalization} offers a straightforward yet effective solution for imbalanced object detection by ignoring gradients for rare categories during network parameter updates, thus helping the model to learn better features for these rare classes.

Replacing the softmax function, Sparsemax \cite{martins2016softmax} produces sparse probability distributions, which allows for more selective attention mechanisms and improves multi-label classification by assigning zero probability to irrelevant classes.





\section{Methodology}
\label{s:method}

\subsection{GBDT models}
Given a dataset 
$\mathcal{D}=\{(\mathbf{x}_1, y_1), (\mathbf{x}_2, y_2), \cdots, (\mathbf{x}_n, y_n)\}$~($\mathbf{x}_i \in \mathbb{R}^m, y_i\in \{0, 1\}$), where $\mathbf{x}_i$ is the $i$th sample and $y_i$ is the corresponding ground-truth label.
Let $l(y, p)$ denote the loss function for binary classification, where $p$ represents the probability for the class labeled as $y = 1$.

In GBDT case, the objective of iteration $t+1$ ($t\ge 0$) is given as follows 
\begin{equation}
\label{e.objfun}
\begin{split}
    \mathcal{L}^{t+1} & = \sum_{i=1}^{n}l(y_i, p^{t+1}_i),\\
                    & = \sum_{i=1}^{n}l(y_i, S(z^{t+1}_i)),\\
                    & = \sum_{i=1}^{n}l(y_i, S(z^{t}_i+\alpha f_{t+1}(\mathbf{x}_i))),
\end{split}
\end{equation}
where $f_{t+1}$ is a new tree, $\alpha$ is the learning rate,
$z^{t}_i = z^{0}_i+\alpha \sum_{j=1}^{t}f_{j}(\mathbf{x}_i)$ is model's raw prediction, and $p^{t+1}_i = S(z^{t+1}_i) = 1/(1+e^{-z^{t+1}_i})$ is obtained by applying the Sigmoid function.

Newton's method is employed to optimize the regularized objective, which has the following formulation:
\begin{equation}
\label{e.objective}
    \mathcal{\widetilde{L}}^{t+1} = \sum\limits_{i=1}^{n}[g^{t}_i f_{t+1}(\mathbf{x}_i)+\frac{1}{2}h^{t}_i f_{t+1}(\mathbf{x}_i)^2] + \Omega(f_{t+1}),\\
\end{equation}
where $\Omega(f_{t+1})$ is a regularization parameter, $g^{t}_i$ is the gradient, and $h^{t}_i$ is the Hessian defined as follows
\begin{equation}
\label{e.binary_grad}
g^{t}_i = \frac{\partial l}{\partial z^{t}_i} = \frac{\partial l}{\partial p^{t}_i}\frac{\partial p^{t}_i}{\partial z^{t}_i},
\end{equation}
\begin{equation}
\label{e.binary_hess}
h^{t}_i = \frac{\partial^2 l}{\partial (z^{t}_i)^2} = \frac{\partial^2 l}{\partial (p^{t}_i)^2}(\frac{\partial p^{t}_i}{\partial z^{t}_i})^2
+\frac{\partial l}{\partial \hat{p}^{t}_i}\frac{\partial^2 p^{t}_i}{\partial (z^{t}_i)^2}.
\end{equation}

Our analysis includes three GBDT models: \cite{chen2016xgboost}, LightGBM \cite{ke2017lightgbm},  XGBoost and SketchBoost \cite{iosipoi2022sketchboost}. XGBoost and LightGBM are two widely-used boosting machine learning algorithms in solving many practical data science problems. SketchBoost is an improvement that can accelerate the training process and extend the scalability of current GBDT models for multioutput problems. We use all three models to evaluate the performances for binary and multi-class tasks. We only employ SketchBoost to assess the performances for multi-label classification since XGBoost and LightGBM do not support this task. It is worth noting that SketchBoost currently only accepts binary loss for multi-label classification.

\subsection{Class-balanced losses for GBDT}
Since not all the class-balanced losses can be used for GBDT, we employ three losses that are suitable for all kinds of classification tasks,  Weighted cross-entropy (WCE) \cite{sun2009classification}, Focal loss (FL) \cite{lin2017focal},  Asymmetric loss (ASL) \cite{ridnik2021asymmetric}. Besides, based on the property of ASL, we newly define two losses, asymmetric weighted cross-entropy (AWE) and asymmetric cross-entropy (ACE). Each loss function's mathematical formulation and implications for addressing class imbalance are discussed in detail.

For simplicity, we only introduce the mathematical formula for binary classification.
Extending the losses to the multi-class case is straightforward and works well.

Suppose $y$ specifies the ground-truth class, and $p$ is the model’s estimated probability for the class with the label $y = 1$. A general form of a binary loss per label, $l$, is given by:
\begin{equation}
    l = -yl_{+}-(1-y)l_{-},
\end{equation}
where $l_{+}$ and $l_{-}$ are the positive and negative loss parts, respectively.

\paragraph{Weighted Cross-entropy}
WCE is obtained by setting $l_{+}$ and $l_{-}$ as:
\begin{equation}
\left\{
\begin{aligned} 
l_+ &= w \log(p),\\
l_- &=  \log(1-p),
\end{aligned}
\right.
\end{equation}
The WCE modifies the standard cross-entropy loss by adding the term  $w~(w>1)$ to the positive loss part,  which enlarges the gradient of the positive class and enables the loss to pay more attention to the this class.

\paragraph{Focal Loss}
FL is obtained by setting $l_{+}$ and $l_{-}$ as:
\begin{equation}
\left\{
\begin{aligned} 
l_+ &= (1-p)^\gamma \log(p) \\
l_- &= p^\gamma \log(1-p)
\end{aligned}
\right.
\end{equation}
where $\gamma $ is a focusing parameter that adjusts the rate at which easy examples are down-weighted.
The FL modifies the standard cross-entropy loss by adding the term  $(1-p)^\gamma$,  which reduces the relative loss for well-classified examples and puts more focus on hard, misclassified examples.

\paragraph{Asymmetric Loss}
ASL is obtained by setting $l_{+}$ and $l_{-}$ as:
\begin{equation}
\left\{
\begin{aligned} 
l_+ &= (1-p)^{\gamma^{+}} \log(p), \\
l_- &= p_m^{\gamma_{-}} \log(1-p_m).
\end{aligned}
\right.
\end{equation}
Here, $p_m$ is the shifted probability defined as $p_m = \max (p-m, 0)$, $m\ge 0$ is the probability margin and is tunable, $\gamma_{+}$ and $\gamma_{-}$ are two focusing parameters, where $\gamma_{-} > \gamma_{+}$.
ASL is a modification of FL that allows us to apply two types of asymmetry for reducing the contribution of easy negative samples to the loss function.

\paragraph{Asymmetric Cross-entropy}
ACE is obtained by setting $l_{+}$ and $l_{-}$ as:
\begin{equation}
\left\{
\begin{aligned} 
l_+ &= \log(p), \\
l_- &= \log(1-p_m).
\end{aligned}
\right.
\end{equation}
ACE has the same idea as ASL; it is a modification of cross-entropy by using the shifted probability in the negative loss part.

\paragraph{Asymmetric Weighted Cross-entropy}
AWE is obtained by setting $l_{+}$ and $l_{-}$ as:
\begin{equation}
\left\{
\begin{aligned} 
l_+ &= w \log(p), \\
l_- &= \log(1-p_m).
\end{aligned}
\right.
\end{equation}
AWE is a modification of WCE by using the shifted probability in the negative loss part.

\subsection{Datasets}
We conduct comprehensive comparisons on 40 datasets, specifically comprising 15 binary, 15 multi-class, and 10 multi-label datasets.
The binary datasets are from the imbalanced-learn \footnote{\url{https://imbalanced-learn.org/}}, the multi-class datasets come from the KEEL \footnote{\url{https://sci2s.ugr.es/keel/datasets.php}}, the multi-label datasets come from the Scikit-multilearn \footnote{\url{http://scikit.ml/}}.
For more details, we refer the reader to \ref{a.data}.

\subsection{Evaluation}
If there is no official train/test split, we randomly split the data into training and test sets with a ratio of 80\% and 20\%, respectively. To tune the hyperparameters of models and losses, we use Optuna \cite{akiba2019optuna} with 100 trials on the training data, each undergoing 5-fold cross-validation where the validation fold is used for early stopping. The best hyperparameters are then evaluated on the test set, calculating multiple performance metrics across 5 folds. For more details about the hyperparameters, please refer to Appendix \ref{a.params}. 

We use the \textbf{F1-score} as the primary evaluation metric. Additionally, we perform a maximum performance comparison by computing the maximum performance of the models using Cross-entropy and comparing it to the maximum performance of the models using class-balanced losses. Baseline Maximum Performance (BMP) is the highest performance metric achieved by any of the three models using Cross-entropy. Class-balanced Maximum Performance (CMP) is the highest performance metric achieved by any of the three models using any of the five class-balanced loss functions. The \textbf{absolute improvement between the BMP and the CMP} indicates how much better (or worse) the best-performing model and loss combination among the new losses is compared to the best-performing model with the baseline loss.

\section{Performance}
\label{s.performance}


\subsection{Binary classification}
Table \ref{T.binary} presents the performance on 15 binary classification datasets.
LightGBM achieves the highest F1-score in 3 out of 15 datasets when using the baseline CE loss. When class-balanced losses are applied, LightGBM shows significant improvement, indicating a strong benefit from these techniques. Notably, WCE consistently provides the most substantial gains.
XGBoost achieves the highest F1-score in only 1 dataset with CE, but class-balanced losses, especially WCE and ASL, markedly improve its performance. SketchBoost also experiences significant enhancements with almost all class-balanced losses. Besides, it obtains the most top results among all the model-loss combinations.
The comparative analysis of LightGBM, XGBoost, and SketchBoost with class-balanced losses reveals several key insights. All three models exhibit strong performance improvements with class-balanced losses, particularly with WCE. 

\begin{figure}[!ht]
    \centering
    \includegraphics[width=0.8\linewidth]{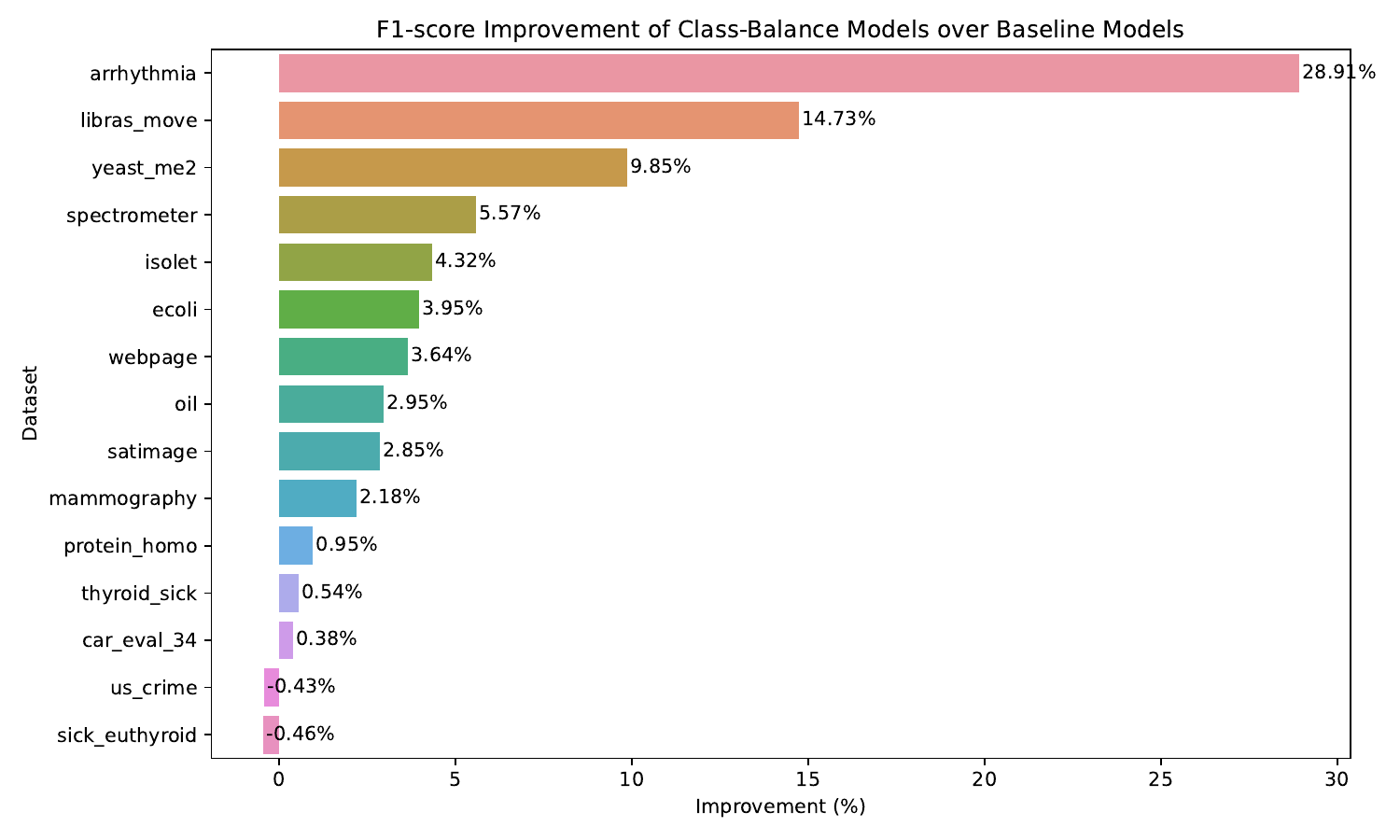}
    \caption{Binary F1-score absolute improvement.}
    \label{f.binaryimprove}
\end{figure}

In Fig.~\ref{f.binaryimprove}, our analysis reveals a predominantly positive trend, with 13 out of 15 datasets exhibiting improvements when class-balance techniques are applied. The magnitude of these improvements varies considerably, spanning from modest gains of 0.38\% to substantial enhancements of up to 28.91\%.
Notably, two datasets (us\_crime and sick\_euthyroid) exhibit slight performance decreases (-0.43\% and -0.46\% respectively), underscoring that while class-balance methods are generally beneficial, they are not universally advantageous and may occasionally lead to overcompensation. This variability highlights the importance of carefully selecting appropriate class-balance techniques for each specific dataset to achieve optimal results.


\begin{table*}[!ht]
\renewcommand\arraystretch{1.4}
\centering
\caption{F1-score performance for each model-loss combination on binary classification datasets. The \textcolor{yellow}{yellow} highlights the best result using LightGBM as the classifier; the \textcolor{cyan}{cyan} highlights the best result using XGBoost as the classifier; the \textcolor{red}{red} highlights the best result using SketchBoost as the classifier; the \underline{\textbf{bold}} indicates the best result among all the model-loss combinations.}

\label{T.binary}
\begin{adjustbox}{width=\textwidth}
\begin{tabular}{l|cccccc|cccccc|cccccc}
\toprule[2pt]

& \multicolumn{6}{c|}{LightGBM} & \multicolumn{6}{c|}{XGBoost} & \multicolumn{6}{c}{SketchBoost} \\
Dataset & CE & WCE & FL & ASL & ACE & AWE & CE & WCE & FL & ASL & ACE & AWE & CE & WCE & FL & ASL & ACE & AWE \\

\midrule[1.5pt]

ecoli &   63.52 &   \cellcolor{yellow}77.11 &   75.13 &    75.44 &    69.40 &    73.27 &   69.83 &    76.52 &   73.50 &    \cellcolor{cyan!30}78.17 &    70.22 &    72.31 &   75.56 &    68.89 &   70.56 &    68.45 &    79.15 &   \cellcolor{red!25}\underline{\textbf{79.51}} \\

satimage &   67.51 &    \cellcolor{yellow}71.91 &   70.71 &    71.44 &  71.84 &  70.54 &    69.05 &   70.02 &    70.81 &   66.45 &    \cellcolor{cyan!30}72.61 &    70.69 &    69.92 &   72.15 &    71.91 &   \cellcolor{red!25}\underline{\textbf{72.77}} &    71.51 &    72.20 \\

sick\_euthyroid &   \cellcolor{yellow}\underline{\textbf{84.84}} &    83.62 &   83.51 &    82.06 &    83.21 &    81.91 &   \cellcolor{cyan!30}84.38 &    83.34 &   83.20 &    79.06 &     84.20 &   81.79 &   \cellcolor{red!25}84.61 & 83.20 & 84.39 & 83.16 & 83.13 & 83.56 \\

spectrometer &   67.24 &    \cellcolor{yellow}71.61 &   70.42 &    67.04 &    70.87 &    68.89 &   75.24 & 74.48 & 72.00 & 77.50 & \cellcolor{cyan!30}\underline{\textbf{80.81}} & 77.14 & 71.18 & 68.74 & 68.74 & 75.18 & 73.33 & \cellcolor{red!25}76.65  \\

car\_eval\_34 &   90.74 &    \cellcolor{yellow}91.19 &   87.19 &    90.75 &    91.16 &    87.53 &   90.74 & \cellcolor{cyan!30}92.20 & 91.47 & 90.95 & 90.69 & 91.74 & 91.98 & \cellcolor{red!25}\underline{\textbf{92.36}} & 92.32 & 90.97 & 91.83 & 89.41 \\

isolet &   84.63 &    \cellcolor{yellow}87.68 &   84.68 &    83.43 &    86.58 &    87.22 & 83.62 & 85.17 & 82.09 & 82.64 & \cellcolor{cyan!30}88.45 & 84.54 & 85.70 & 88.08 & 89.00 & 88.34 & \cellcolor{red!25}\underline{\textbf{90.02}} & 89.58 \\

us\_crime & \cellcolor{yellow}\underline{\textbf{54.94}} & 53.59 & 54.23 & 54.51 & 53.01 & 53.40 & 51.62 & \cellcolor{cyan!30}53.66 & 53.50 & 51.77 & 51.90 & 52.32 & 52.24 & \cellcolor{red!25}53.99 & 48.67 & 51.62 & 52.15 & 53.97 \\

libras\_move & 74.67 & 74.31 & 61.33 & \cellcolor{yellow}76.18 & 72.00 & 67.47 & 68.67 & 73.21 & 68.67 & \cellcolor{cyan!30}83.58 & 79.52 & 78.06 & 72.00 & 79.52 & 75.39 & \cellcolor{red!25}\underline{\textbf{89.39}} & 87.21 & 81.45 \\

thyroid\_sick & 90.54 & \cellcolor{yellow}91.36 & 90.78 & 90.89 & 90.17 & 89.89 & 90.96 & 91.60 & 91.36 & 88.24 & 91.06 & \cellcolor{cyan!30}91.71 & 92.68 & \cellcolor{red!25}\underline{\textbf{93.23}} & 91.96 & 90.62 & 91.28 & 92.77 \\

arrhythmia & 55.33 & 84.36 & 56.67 & \cellcolor{yellow}\underline{\textbf{90.91}} & \cellcolor{yellow}\underline{\textbf{90.91}} & 81.70 & 62.00 & \cellcolor{cyan!30}82.18 & 63.33 & 74.67 & 74.18 & 79.52 & 61.33 & 69.70 & 66.00 & \cellcolor{red!25}\underline{\textbf{90.91}} & 76.85 & 84.36 \\

oil & 32.26 & 36.12 & 29.39 & 36.00 & 41.90 & \cellcolor{yellow}42.24 & 31.76 & 37.80 & 37.09 & \cellcolor{cyan!30}\underline{\textbf{43.09}} & 38.19 & 36.62 & 40.14 & 30.89 & 24.67 & 38.10 & \cellcolor{red!25}42.34 & 37.72 \\

yeast\_me2 & 12.38 & 30.32 & 18.52 & 32.38 & 16.21 & \cellcolor{yellow}\underline{\textbf{34.04}} & 24.18 & 29.32 & 19.88 & \cellcolor{cyan!30}33.77 & 25.95 & 29.18 & 23.45 & 21.34 & 12.57 & 17.89 & \cellcolor{red!25}23.96 & 18.69 \\

webpage & 73.72 & \cellcolor{yellow}75.44 & 73.46 & 74.88 & 74.16 & 74.86 & 74.67 & 76.61 & 76.00 & 78.75 & 78.44 & \cellcolor{cyan!30}80.88 & 77.62 & 77.10 & 74.64 & 80.48 & 79.88 & \cellcolor{red!25}\underline{\textbf{81.26}} \\

mammography & \cellcolor{yellow}69.96 & 67.54 & 66.29 & 68.84 & 66.18 & 66.47 & 67.25 & 67.76 & 68.54 & 67.99 & 59.77 & \cellcolor{cyan!30}70.50 & 67.13 & 68.10 & 67.91 & \cellcolor{red!25}\underline{\textbf{72.14}} & 69.62 & 71.13 \\

protein\_homo & 86.25 & 87.49 & \cellcolor{yellow}87.83 & 87.30 & 87.29 & 85.29 & 86.98 & \cellcolor{cyan!30}87.32 & 86.94 & 85.05 & 87.30 & 83.92 & 87.49 & 86.80 & 87.92 & 88.22 & 87.85 & \cellcolor{red!25}\underline{\textbf{88.44}} \\

\bottomrule[2pt]
\end{tabular}
\end{adjustbox}
\end{table*}

\subsection{Multi-class classification}

Table \ref{T.mcc} presents the performance on 15 multi-class classification datasets.
Unlike in binary classification, LightGBM achieves the highest F1-score in 6 out of 15 datasets when using the baseline CE loss in multi-class settings. While class-balanced losses do provide improvements, their overall impact is less pronounced compared to binary datasets, likely due to the increased complexity of multi-class datasets.
XGBoost achieves the highest F1-score in only 3 datasets with CE, but class-balanced losses, especially FL, markedly improve its performance. SketchBoost experiences significant enhancements with almost all class-balanced losses and secures the most top results among all model-loss combinations, thanks to its efficient handling of multioutput tasks.
All three models consistently exhibit strong performance improvements with class-balanced losses in multi-class settings. However, there is a subtle difference compared to binary classification. Notably, the best results are all achieved by models using class-balanced losses. This suggests that multi-class imbalanced classification can benefit more from these techniques.

\begin{figure}[!ht]
    \centering
    \includegraphics[width=0.8\linewidth]{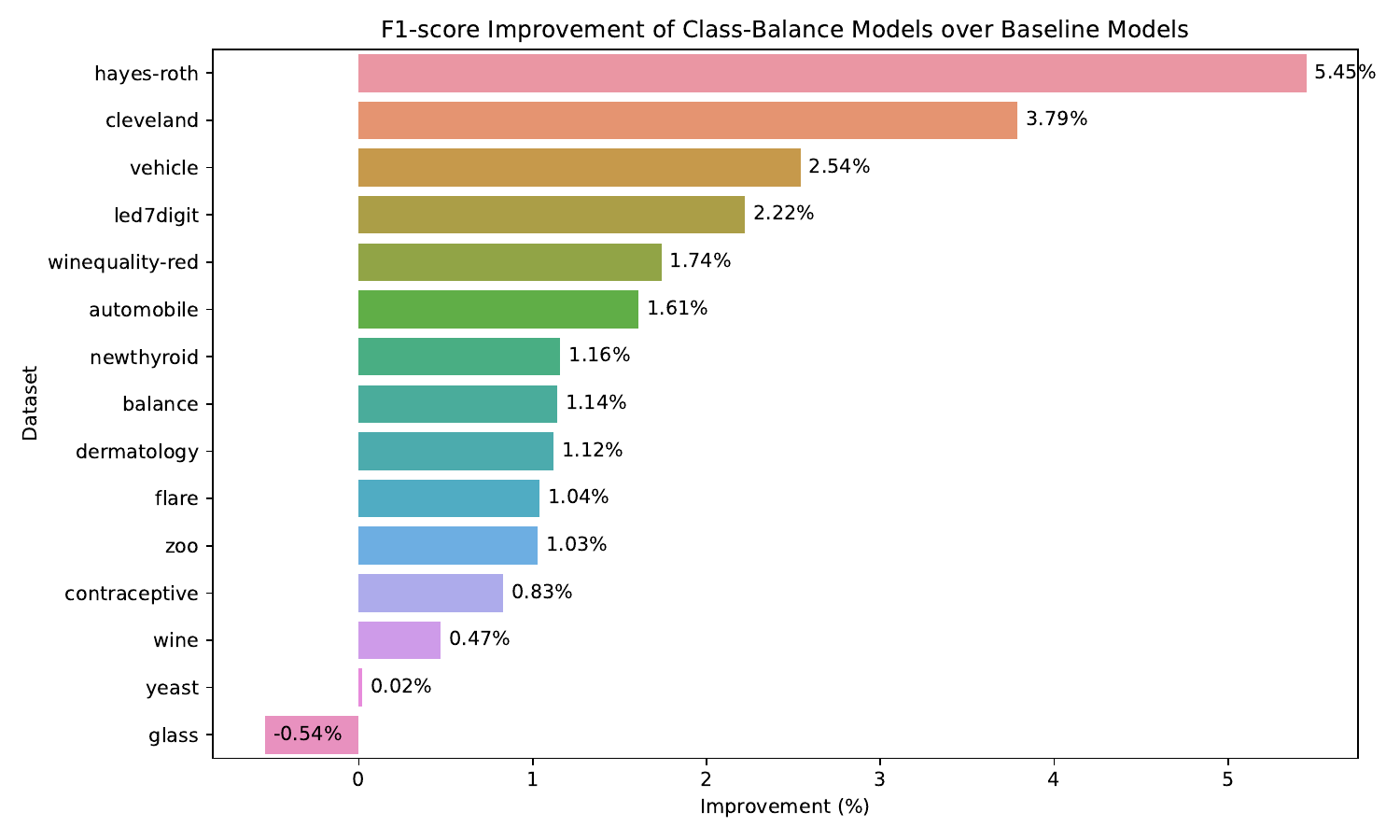}
    \caption{Multi-class F1-score absolute improvement.}
    \label{f.mccimprove}
\end{figure}

In Fig.~\ref{f.mccimprove}, the results show that 14 out of 15 datasets exhibit improvements when class-balance techniques are applied, with enhancements ranging from a modest 0.02\% to a more substantial 5.45\%.
However, the magnitude of improvements in multi-class classification is generally smaller compared to binary classification. For instance, while the majority of datasets in binary classification showed improvements above 2\%, most multi-class datasets exhibit improvements below 2\%. This difference suggests that class imbalance may have a more pronounced effect in binary classification scenarios, or that addressing multi-class imbalance is a more complex challenge.

\begin{table*}[!ht]
\renewcommand\arraystretch{1.4}
\centering
\caption{F1-score performance for each model-loss combination on multi-class classification datasets. The \textcolor{yellow}{yellow} highlights the best result using LightGBM as the classifier; the \textcolor{cyan}{cyan} highlights the best result using XGBoost as the classifier; the \textcolor{red}{red} highlights the best result using SketchBoost as the classifier; the \underline{\textbf{bold}} indicates the best result among all the model-loss combinations.}
\label{T.mcc}
\begin{adjustbox}{width=\textwidth}
\begin{tabular}{l|cccccc|cccccc|cccccc}
\toprule[2pt] 

& \multicolumn{6}{c|}{LightGBM} & \multicolumn{6}{c|}{XGBoost} & \multicolumn{6}{c}{SketchBoost} \\
Dataset & CE & WCE & FL & ASL & ACE & AWE & CE & WCE & FL & ASL & ACE & AWE & CE & WCE & FL & ASL & ACE & AWE \\

\midrule[1.5pt]

wine & 92.82 & 92.02 & 90.99 & \cellcolor{yellow}93.30 & 92.82 & 92.00 & \cellcolor{cyan!30}93.74 & 91.13 & 92.82 & 91.01 & 91.50 & 93.30 & 94.65 & 93.79 & 94.67 & 93.35 & \cellcolor{red!25}\underline{\textbf{95.12}} & 92.48 \\

zoo & 89.85 & 84.99 & 89.75 & 89.94 & \cellcolor{yellow}90.59 & 77.98 & \cellcolor{cyan!30}93.59 & 91.03 & 91.03 & 87.44 & 88.97 & 91.03 & 93.56 & \cellcolor{red!25}\underline{\textbf{94.62}} & 92.44 & 94.50 & 91.77 & 92.77 \\

glass & \cellcolor{yellow}74.77 & 73.04 & 71.67 & 73.53 & 71.03 & 72.50 & 72.19 & 77.43 & 68.72 & 74.77 & 76.89 & \cellcolor{cyan!30}78.77 & \cellcolor{red!25}\underline{\textbf{79.31}} & 73.41 & 76.27 & 78.25 & 78.10 & 76.70 \\

dermatology & 96.21 & 93.20 & 90.44 & \cellcolor{yellow}96.45 & 96.27 & 91.39 & 95.31 & \cellcolor{cyan!30}95.97 & 95.53 & 95.32 & 94.96 & 94.78 & 96.23 & 95.53 & 97.11 & 96.66 & \cellcolor{red!25}\underline{\textbf{97.35}} & 96.68 \\

led7digit & \cellcolor{yellow}79.30 & 76.44 & 78.54 & 78.43 & 77.36 & 77.17 & 79.15 & 75.99 & 77.43 & 78.06 & 79.16 & \cellcolor{cyan!30}79.85 & 78.80 & 80.19 & 80.02 & \cellcolor{red!25}\underline{\textbf{81.52}} & 77.32 & 80.34 \\

hayes-roth & 67.07 & 78.32 & 74.60 & 79.22 & \cellcolor{yellow}79.52 & 78.82 & 81.41 & 84.35 & 82.76 & 77.43 & 76.53 & \cellcolor{cyan!30}\underline{\textbf{86.86}} & 77.52 & 78.71 & 77.09 & 79.56 & 82.90 & \cellcolor{red!25}83.34 \\

yeast & \cellcolor{yellow}57.76 & 55.32 & 56.91 & 55.56 & 55.63 & 55.83 & \cellcolor{cyan!30}59.15 & 58.52 & 58.37 & 57.57 & 58.03 & 58.21 & 56.39 & 57.49 & 58.14 & 58.75 & 58.67 & \cellcolor{red!25}\underline{\textbf{59.17}} \\

automobile & 82.06 & 83.35 & \cellcolor{yellow}85.30 & 77.15 & 78.76 & 73.80 & 84.77 & 79.04 & \cellcolor{cyan!30}\underline{\textbf{86.38}} & 78.53 & 72.99 & 67.23 & 83.60 & 79.56 & 84.88 & 82.26 & \cellcolor{red!25}86.24 & 80.41 \\

vehicle & 76.35 & \cellcolor{yellow}\underline{\textbf{79.78}} & 75.07 & 74.29 & 74.65 & 75.79 & 77.24 & 76.92 & \cellcolor{cyan!30}78.00 & 76.16 & 75.22 & 77.19 & 75.89 & 76.86 & \cellcolor{red!25}77.54 & 76.83 & 77.42 & 77.49 \\

newthyroid & 92.40 & 91.97 & \cellcolor{yellow}93.13 & 91.99 & 92.53 & 90.91 & 91.41 & 91.08 & 90.47 & 89.10 & 90.65 & \cellcolor{cyan!30}91.44 & 94.42 & \cellcolor{red!25}\underline{\textbf{95.58}} & 93.99 & 92.00 & 94.73 & 93.09 \\

balance & \cellcolor{yellow}88.32 & 86.23 & 87.58 & 86.36 & 86.92 & 86.36 & 88.40 & 87.35 & \cellcolor{cyan!30}90.19 & 89.09 & 89.18 & 87.96 & 90.51 & 88.22 & 89.14 & \cellcolor{red!25}\underline{\textbf{91.65}} & 89.21 & 89.00 \\

cleveland & \cellcolor{yellow}54.98 & 50.82 & 48.33 & 54.04 & 49.83 & 50.85 & 53.84 & 53.05 & \cellcolor{cyan!30}54.76 & 47.88 & 50.20 & 42.64 & 49.98 & 53.13 & 55.22 & 53.34 & 54.08 & \cellcolor{red!25}\underline{\textbf{58.77}} \\

flare & \cellcolor{yellow}72.07 & 71.49 & 71.47 & 71.68 & 71.89 & 71.13 & 71.22 & 70.60 & \cellcolor{cyan!30}71.38 & 71.00 & 68.80 & 68.97 & 71.69 & 71.61 & 72.50 & \cellcolor{red!25}\underline{\textbf{73.11}} & 71.91 & 71.47 \\

contraceptive & 55.78 & \cellcolor{yellow}56.62 & 56.22 & 55.49 & 55.00 & 56.12 & 56.34 & 55.96 & 56.83 & 56.63 & 56.64 & \cellcolor{cyan!30}56.90 & 54.57 & 57.01 & \cellcolor{red!25}\underline{\textbf{57.17}} & 54.77 & 55.01 & 55.42 \\

winequality-red & 57.83 & \cellcolor{yellow}59.29 & 57.17 & 58.69 & 57.69 & 58.84 & 60.32 & 58.47 & \cellcolor{cyan!30}\underline{\textbf{62.06}} & 61.59 & 58.22 & 61.79 & 60.21 & 59.38 & 59.98 & \cellcolor{red!25}60.66 & 60.03 & 60.60 \\

\bottomrule[2pt]
\end{tabular}
\end{adjustbox}
\end{table*}

\subsection{Multi-label classification}

Table \ref{T.mlc} shows the application of class-balance loss functions to multi-label classification tasks demonstrates a notably more substantial impact compared to both binary and multi-class scenarios.

\begin{figure}[!ht]
    \centering
    \includegraphics[width=0.8\linewidth]{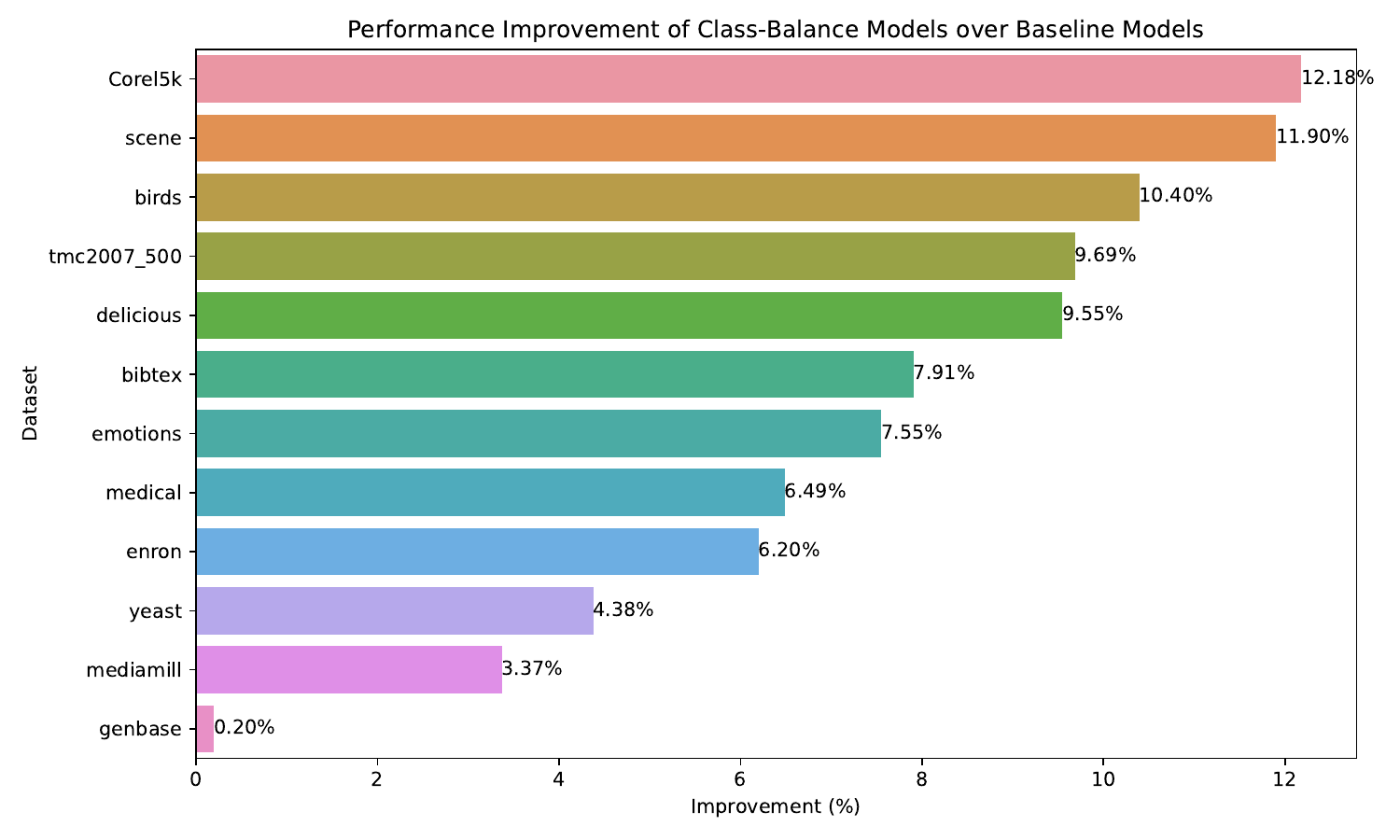}
    \caption{Multi-label F1-score absolute improvement.}
    \label{f.mlcimprove}
\end{figure}

All 10 datasets in the multi-label experiment show positive improvements (Fig.~\ref{f.mlcimprove}), with enhancements ranging from 0.20\% to an impressive 12.18\%.
The most significant improvements are observed in datasets such as Corel5k (12.18\%), scene (11.90\%), and birds (10.40\%). These improvements are considerably larger than those seen in both binary and multi-class classifications. For context, the highest improvement in binary classification is 28.91\% for the arrhythmia dataset, while in multi-class, the top improvement is 5.45\% for hayes-roth. The fact that three multi-label datasets show improvements above 10\% is particularly noteworthy.
Moreover, the overall magnitude of improvements in multi-label classification is generally higher than in both binary and multi-class scenarios. While binary classification has several datasets with improvements between 2-5\%, and multi-class mostly has improvements below 2\%, multi-label classification shows a majority of datasets (7 out of 10) with improvements above 5\%. This suggests that class-balance techniques may be especially effective in addressing the complexities of multi-label imbalance.

\begin{table*}[!ht]
\renewcommand\arraystretch{1.3}
\footnotesize
\centering
\caption{F1-score performance for each model-loss combination on multi-label classification datasets. The \textcolor{red}{red} highlights the best result using SketchBoost as the classifier; the \underline{\textbf{bold}} indicates the best result.}
\label{T.mlc}

\begin{adjustbox}{width=0.6\textwidth}
\begin{tabular}{l|cccccc}
\toprule[2pt] 

&  \multicolumn{6}{c}{SketchBoost} \\
Dataset & CE & WCE & FL & ASL & ACE & AWE  \\

\midrule[1.5pt]

Corel5k & 10.49 & 20.03 & 15.08 & \cellcolor{red!25}\underline{\textbf{22.67}} & 21.07 & 21.31 \\

bibtex & 35.47 & 40.42 & 36.66 & 43.32 & \cellcolor{red!25}\underline{\textbf{43.38}} & 42.72 \\

birds & 11.59 & 16.78 & 17.29 & \cellcolor{red!25}\underline{\textbf{21.98}} & 21.09 & 21.88 \\

emotions & 59.24 & 64.53 & 59.22 & 64.87 & 65.54 & \cellcolor{red!25}\underline{\textbf{66.79}} \\

enron & 53.30 & \cellcolor{red!25}\underline{\textbf{59.50}} & 54.00 & 58.21 & 59.25 & 59.46 \\

mediamill & 55.65 & 57.89 & 56.06 & 57.13 & 58.37 & \cellcolor{red!25}\underline{\textbf{59.01}} \\

medical & 69.27 & 74.21 & 72.08 & \cellcolor{red!25}\underline{\textbf{75.76}} & 74.78 & 74.39 \\

scene & 65.78 & 74.38 & 69.25 & \cellcolor{red!25}\underline{\textbf{77.69}} & 75.96 & 74.97 \\

tmc2007\_500 & 80.49 & 83.11 & 81.24 & 83.07 & 83.16 & \cellcolor{red!25}\underline{\textbf{90.18}} \\

yeast & 61.59 & 65.53 & 62.01 & \cellcolor{red!25}\underline{\textbf{65.97}} & 65.68 & 65.22 \\

\bottomrule[2pt]
\end{tabular}
\end{adjustbox}
\end{table*}


\subsection{Summary}
The above results demonstrate that these techniques generally improve model performance on imbalanced datasets, with varying degrees of effectiveness across the different task types. Different models and class-balanced techniques perform best on different datasets, highlighting the importance of selecting appropriate methods for each specific case.

The effectiveness of class-balanced techniques appears to increase with task complexity, showing the largest improvements in multi-label classification. This trend suggests that these methods are particularly adept at addressing the intricate imbalances present in multi-label problems.

Across all task types and models, it is hard to say which GBDT model and which class-balanced loss is the best choice. The optimal class-balanced technique varies by dataset, underscoring the importance of careful method selection. 

These findings highlight the potential of class-balanced loss functions in improving GBDT model performance on imbalanced datasets, especially in complex classification scenarios. The variation in results across different GBDT implementations emphasizes the importance of model selection in addition to the choice of class-balanced technique.

\section{Python package implementations}
In our implementation, XGBoost and LightGBM share the same implementations for loss functions in binary classification but differ in their multi-class classification strategies. SketchBoost maintains a unified implementation for custom loss functions in binary and multi-label classification but adopts a separate method for multi-class classification. 

The class-balanced loss functions implemented in our Python package  are summarized in Table \ref{T.loss}. A '$\checkmark$' indicates that the loss function has been included in the package, while a '-' indicates that the loss function is not suitable for the task.

\begin{table*}[!ht]
\footnotesize
\centering
\caption{Class-balanced loss in Python package}
\label{T.loss}
\begin{adjustbox}{width=0.9\textwidth}
\begin{tabular}{lccc}
\toprule[1.5pt]
       & Binary & Multi-class & Multi-label \\
\midrule[1pt]
Weighted cross-entropy \cite{sun2009classification}& $\checkmark$ & $\checkmark$ & $\checkmark$ \\
\hline
Focal loss \cite{lin2017focal}& $\checkmark$ & $\checkmark$ & $\checkmark$ \\
\hline
Asymmetric loss \cite{ridnik2021asymmetric}& $\checkmark$ & $\checkmark$ & $\checkmark$ \\
\hline
Asymmetric cross-entropy & $\checkmark$ & $\checkmark$ & $\checkmark$ \\
\hline
Asymmetric weighted cross-entropy & $\checkmark$ & $\checkmark$ & $\checkmark$ \\
\hline
Class-balanced cross-entropy \cite{cui2019class}& $\checkmark$ & $\checkmark$ & - \\

\bottomrule[1.5pt]
\end{tabular}
\end{adjustbox}
\end{table*}

\section{Conclusions}
\label{s:conclu}

In this study, we have presented a thorough investigation into the adaptation of class-balanced loss functions for GBDT algorithms in tabular classification tasks. Our comprehensive experiments across diverse datasets have demonstrated the efficacy of these loss functions in addressing class imbalance issues, a common challenge in real-world machine learning applications.

In addition, the introduction of our Python package represents a significant step towards making class-balanced loss functions more accessible and easier to implement within GBDT frameworks. This tool has the potential to accelerate research and improve practical applications in the field.

Our findings underscore the potential of class-balanced loss functions to improve the performance of GBDT models on imbalanced datasets, addressing a critical challenge in machine learning. Future research directions include exploring the synergy between class-balanced losses and other techniques such as sampling methods, extending their application to multi-class and multi-label classification problems, and evaluating their effectiveness in other tree-based ensemble methods. 






\appendix
\section{Datasets Description}
\label{a.data}
Table. \ref{T.data} lists the datasets used in this paper. \#S means the sample number; \#F indicates the feature number; \#C gives the number of class; IR is the  imbalanced ratio. For binary and multi-class classification, it is the ratio of the most frequent class to the least frequent class. For multi-label classification, it is the ratio of the most frequent label to the least frequent label.

\begin{table*}[!ht]
\footnotesize
\centering
\caption{Dataset description}
\label{T.data}
\begin{adjustbox}{width=0.9\textwidth}
\begin{tabular}{lcccc|lcccc}
\toprule[1.5pt]
Name & \#S & \#F & \#C & IR & Name & \#S & \#F & \#C & IR\\
\midrule[1pt]

\multicolumn{10}{c}{Binary} \\
\hline
ecoli & 336 & 7 & 2 & 8.6 & satimage & 6435 & 36 & 2 & 9.3\\
\hline
sick\_euthyroid & 3163 & 42 & 2 & 9.8 & spectrometer & 531 & 93 & 2 & 11\\
\hline
car\_eval\_34 & 1728 & 21 & 2 & 12 & isolet & 7797 & 617 & 2 & 12\\
\hline
us\_crime & 1994 & 100 & 2 & 12 & libras\_move & 360 & 90 & 2 & 14\\
\hline
thyroid\_sick & 3772 & 52 & 2 & 15 & arrhythmia & 452 & 278 & 2 & 17\\
\hline
oil & 937 & 49 & 2 & 22 & yeast\_me2 & 1484 & 8 & 2 & 28\\
\hline
webpage & 34780 & 300 & 2 & 33 & mammography & 11183 & 6 & 2 & 42 \\
\hline
protein\_homo & 145751 & 74 & 2 & 111 &&&&\\
\midrule[1pt]

\multicolumn{10}{c}{Multi-class} \\
\hline
wine & 173 & 13 & 3 & 1.5 & zoo & 101 & 16 & 7 & 10.3\\
\hline
glass & 214 & 9 & 6 & 8.43 & dermatology & 358 & 34 & 6 & 5.5\\
\hline
led7digit & 500 & 7 & 10 & 1.5 & hayes-roth & 160 & 4 & 3 & 2.1\\
\hline
yeast & 1484 & 8 & 10 & 91.8 & automobile & 159 & 64 & 6 & 16\\
\hline
vehicle & 846 & 18 & 4 & 1.1 & newthyroid & 215 & 5 & 3 & 5\\
\hline
balance & 625 & 4 & 3 & 5.9& cleveland & 297 & 13 & 5 & 12.3\\
\hline
flare & 1066 & 19 & 6 & 7.7 & contraceptive & 1473 & 9 & 3 & 39.2\\
\hline
winequality-red & 1599 & 11 & 6 & 67.6 &&&&\\
\midrule[1pt]

\multicolumn{10}{c}{Multi-label} \\
\hline
Corel5k & 4500 & 499 & 374 & 1004 & bibtex & 4880 & 1836 & 159 & 24.7\\
\hline
birds & 322 & 260 & 19 & 16 & emotions & 391 & 72 & 6 & 1.9\\
\hline
enron & 1123 & 1001 & 53 & 600 & mediamill & 30993 & 120 & 101 & 6017.8\\
\hline
medical & 333 & 1449 & 45 & 98 & scene & 1211 & 294 & 6 & 1.7 \\
\hline
tmc2007\_500 & 21519 & 500 & 2 & 42.4 & yeast & 1500 & 103 & 14 & 53.7\\

\bottomrule[1.5pt]
\end{tabular}
\end{adjustbox}
\end{table*}

\section{Optimization of hyperparameters}
\label{a.params}
Table. \ref{T.hyperparam} lists the search range of hyperparameters. 

\begin{table*}[!ht]
\footnotesize
\centering
\caption{Hyperparameters for GBDT models}
\label{T.hyperparam}
\begin{adjustbox}{width=\textwidth}
\begin{tabular}{llll}
\toprule[2pt]
HyperParameters & Range & HyperParameters & Range \\
\midrule[1.5pt]

\multicolumn{4}{c}{LightGBM} \\
\hline
$iterations$ & 1000 & $early\_stopping\_round$ & 50\\
\hline
$num\_leaves$ & LogUniformInt [8, 64] & $reg\_alpha$ & LogUniform [1e-4, 2]\\
\hline
$reg\_lambda$ & LogUniform [1e-4, 2] & $learning\_rate$ & LogUniform [0.01, 1.0] \\
\midrule[1.5pt]

\multicolumn{4}{c}{XGBoost} \\
\hline
$num\_boost\_round$ & 1000  & $early\_stopping\_rounds$ & 50  \\
\hline
$max\_depth$ & LogUniformInt [2, 10] & $reg\_alpha$ & LogUniform [1e-4, 1]  \\
\hline
$reg\_lambda$ & LogUniform [1e-4, 5] & $eta$ & LogUniform [1e-3, 1]  \\
\midrule[1.5pt]

\multicolumn{4}{c}{SketchBoost} \\
\hline
$ntrees$ & 1000 & $early\_stopping\_rounds$ & 50 \\
\hline
$max\_depth$ & LogUniformInt [2, 10] & $lambda\_{l_2}$ & LogUniform [1e-4, 2]\\
\hline
$learning\_rate$ & LogUniform [0.01, 1] & $max\_bin$ & LogUniformInt [64, 256]\\
\hline
$subsample$ & LogUniform [0.05, 1] &&\\

\midrule[1.5pt]
\multicolumn{4}{c}{Class-balanced loss} \\
\hline
$WCE$ & $w:$~[2, 3, 5] & & \\
\hline
$FL$  & $\gamma:$~[0.5, 1, 2] && \\
\hline
$ASL$ & $\gamma^{+}:$~[0.0, 0.1] & $\gamma^{-}:$~[0.5, 1, 2] & $m:$~[0.05, 0.2]\\
\hline
$ACE$ & $m:$~[0.05, 0.2] && \\
\hline
$AWE$ & $w:$~[2, 3, 5] & $m:$~[0.05, 0.2] &\\

\bottomrule[2pt]
\end{tabular}
\end{adjustbox}
\end{table*}

\bibliographystyle{elsarticle-num} 
\bibliography{references}

\end{document}